\documentclass[conference]{IEEEtran}

 \usepackage[dvipsnames,table,xcdraw]{xcolor}
\usepackage{pgfplots}
\usepackage{pgfplotstable}
\usepackage{booktabs}
\usepackage{colortbl}
\usepackage{float}
\usepackage{url}
\usepackage{booktabs}
\usepackage{pdfpages}
\usepackage{svg}
\usepackage{hyperref}
\usepackage{subcaption}
\usepackage{multirow}
\usepackage[utf8]{inputenc}
\usepackage{authblk}
\usepackage[T1]{fontenc}
\usepackage{amsmath,amssymb}
\usepackage{paralist}
\usepackage{latexsym}
\usepackage{wasysym}
\usepackage{graphicx}
\usepackage{MnSymbol}
\usepackage{stackengine}
\usepackage{mdframed}
\usepackage{comment}
\usepackage[ruled,linesnumbered,commentsnumbered,vlined]{algorithm2e}
\usepackage{enumitem}
\usepackage{multicol}
\def\baselinestretch{0.99}
\usepackage{menukeys}

\usepackage{amsmath}
\usepackage{tikz-qtree}

\usepackage[listings,skins,breakable]{tcolorbox}

\usetikzlibrary{shapes} 
\usetikzlibrary{fit,backgrounds}
\def\addsquare#1{\tikz\node[minimum size=0.7cm,draw]{#1};} 

\SetAlFnt{\scriptsize}

\DeclareMathAlphabet{\mathcal}{OMS}{cmsy}{m}{n}

\renewcommand{\figurename}{Fig.}
\renewcommand{\tablename}{Tab.}

\newcommand{\noleftdelimiter}{\left.\kern-\nulldelimiterspace}

\newcommand{\autour}[1]{\tikz[baseline=(X.base)]\node [draw=black,fill=cyan!20,thick,rectangle,rounded corners=3pt,text depth=0pt] (X) {#1};}

\makeatletter
\def\ps@IEEEtitlepagestyle{%
  \def\@oddfoot{\mycopyrightnotice}%
  \def\@evenfoot{}%
}
\def\mycopyrightnotice{%
  {\begin{minipage}{\textwidth}
  \footnotesize \copyright 2021 IEEE. Personal use of this material is permitted. Permission from IEEE must be obtained for all other uses, in any current or future media, including reprinting\slash republishing this material for advertising or promotional purposes, creating new collective works, for resale or redistribution to servers or lists, or reuse of any copyrighted component of this work in other works.
  \end{minipage}
  }
  \gdef\mycopyrightnotice{}
}

\begin{document}

\title{Fine-Grained Causality Extraction From Natural Language Requirements Using Recursive Neural Tensor Networks}

\author[1]{Jannik Fischbach}
\author[2]{Tobias Springer}
\author[3]{Julian Frattini}
\author[1]{Henning Femmer}
\author[4]{\\ Andreas Vogelsang}
\author[3,5]{Daniel Mendez}

\affil[1]{\footnotesize Qualicen GmbH, Germany, \{firstname.lastname\}@qualicen.de}
\affil[2]{\footnotesize Technical University of Munich, Germany, 
tobias.springer@tum.de}
\affil[3]{\footnotesize Blekinge Institute of Technology, Sweden, \{firstname.lastname\}@bth.se}
\affil[4]{\footnotesize University of Cologne, Germany, vogelsang@cs.uni-koeln.de}
\affil[5]{\footnotesize fortiss GmbH, Germany, mendez@fortiss.org}

\maketitle

\begin{abstract}
[Context:] Causal relations (e.g., \textit{If A, then B}) are prevalent in functional requirements. For various applications of AI4RE, e.g., the automatic derivation of suitable test cases from requirements, automatically extracting such causal statements are a basic necessity. 
[Problem:] We lack an approach that is able to extract causal relations from natural language requirements in fine-grained form. Specifically, existing approaches do not consider the combinatorics between causes and effects. They also do not allow to split causes and effects into more granular text fragments (e.g., variable and condition), making the extracted relations unsuitable for automatic test case derivation.
[Objective \& Contributions:] We address this research gap and make the following contributions: First, we present the \emph{Causality Treebank}, which is the first corpus of fully labeled binary parse trees representing the composition of 1,571 causal requirements. Second, we propose a fine-grained causality extractor based on Recursive Neural Tensor Networks. Our approach is capable of recovering the composition of causal statements written in natural language and achieves a F1 score of 74\% in the evaluation on the \emph{Causality Treebank}. Third, we disclose our open data sets as well as our code to foster the discourse on the automatic extraction of causality in the RE community.
\end{abstract}

\section{Introduction}

\textbf{Motivation.} Natural language (NL) is the most common notation for expressing requirements~\cite{Mich04,Kassab14}. Functional requirements describe the expected system behavior often based on causal relations (e.g., \textit{If the user enters an incorrect password, the system shows an error message}). In previous studies~\cite{fischbachREFSQ,fischbachICST} we were able to show that such conditional statements are prevalent in both traditional RE documents and agile RE artifacts, such as acceptance criteria. Automatically extracting causal relations from NL requirements and utilizing their embedded logical knowledge supports two use cases~\cite{fischbachRENEXT}: First, the automatic derivation of test cases from requirements. Second, the detection of dependencies between requirements. 

\textbf{Related Work \& Research Gap.} We lack an approach that is able to extract causal relations from NL requirements in fine-grained form. More specifically, the current state of research has the following gaps:


\begin{enumerate}[label=\bfseries P\arabic*:,leftmargin=*,labelindent=0em]
    \item Rule-based systems~\cite{Garcia97, khoo98, puente} are highly dependent on hand-crafted patterns. For this reason, more recent approaches~\cite{Asghar16,yang2021survey} apply Deep Learning (DL) in order to automatically extract useful features from raw text. 
    However, the existing approaches~\cite{Ponti17,canasai17,Jin2020} have been trained on corpora not originating from software engineering (e.g., BBC news article set~\cite{Greene06}) and are therefore difficult to utilize for RE purposes. Since RE documents often exhibit a specific vocabulary, we require an approach that is trained on RE data~\cite{ferrari17}.

\item Neither the code nor any demos are publicly available for the existing approaches. Hence, they can not be used without great re-implementation efforts.

\item The greatest issue, however, lies in the way causes and effects are extracted. Let us assume the following requirement \keys{REQ 1}: \textit{If A is true and B is false, then C shall occur}. Some approaches~\cite{chang05,rink10} extract causes and effects only on word level (i.e., $\boldsymbol{\mathsf{cause_1}}$: \textit{A}, $\boldsymbol{\mathsf{cause_2}}$: \textit{B}, $\boldsymbol{\mathsf{effect_1}}$: \textit{C}). Consequently, valuable information about the causal relation is lost (e.g., the conditions of \textit{A}, \textit{B} and \textit{C} are ignored). Others~\cite{dasgupta18,li19} manage to extract causes and effects at phrase level (i.e., $\boldsymbol{\mathsf{cause_1}}$: \textit{A is true}, $\boldsymbol{\mathsf{cause_2}}$: \textit{B is false}, $\boldsymbol{\mathsf{effect_1}}$: \textit{C shall occur}). However, these extracted text fragments are not fine-grained enough for the mentioned use cases. In order to derive test cases, causes and effects must be further decomposed into variable and condition (i.e., $\boldsymbol{\mathsf{cause\_variable_1}}$: \textit{A}, $\boldsymbol{\mathsf{cause\_condition_1}}$: \textit{is true}). In addition, we need to understand the combinatorics between the causes and effects and extract the relation accordingly: \keys{REQ 1} states that A and B are supposed to occur together before C shall occur. 
\end{enumerate}

\textbf{Objective.} Our goal is to develop a new approach that addresses the presented issues and extracts causal relations from NL requirements with reasonable performance. For this purpose, we implement a Recursive Neural Tensor Network (RNTN)~\cite{socher13} in order to extract causes and effects in fine-grained form.

\textbf{Contributions}: In this paper, we make the following contributions (C):

\begin{enumerate}[label=\bfseries C\arabic*:,leftmargin=*,labelindent=0em]
    \item We present the \emph{Causality Treebank} which is the first corpus of fully labeled binary parse trees representing the composition of causal relations in NL requirements. Specifically, we annotated the composition of 1,571 causal requirements on the basis of 27 different segment labels (e.g., causes, conditions, conjunctions).
    \item We present a fine-grained causality extractor built on a RNTN. We train and empirically evaluate our approach based on the \emph{Causality Treebank} and achieve a F1 score of 74\% across all segments.
    \item To strengthen transparency and facilitate replication, we disclose our code and annotated data sets.\footnote{\label{note1}Our code and annotated data sets can be found at \url{https://github.com/springto/Fine-Grained-Causality-Extraction-From-NL-Requirements}.}
\end{enumerate}

\section{Recursive Neural Tensor Networks}
This section motivates the usage of Recursive Neural Tensor Networks (RNTN)~\cite{socher13} for causality extraction and provides an overview of its functionality.

\subsection{Why do we use RNTN?}
 A RNTN is based on the idea that natural language can be understood as a recursive structure~\cite{socher11}. For example, the syntax of a sentence is recursively structured, with noun phrases containing relative phases, which in turn contain further noun phrases, and so on. A RNTN is capable of recovering this recursive structure and helps to better understand the composition of a sentence. We argue that a causal relation also represents a recursive structure as it consists of causes and effects, which in case of conjunctions and disjunctions consists of further causes and effects, and so on~\cite{fischbachRENEXT}. This results in a tree-like structure of cause and effect nodes forming the full sentence. By recovering this tree-like structure, we do not lose the combinatorics between the causes and effects which allows us to entirely extract the causal relation. Furthermore, it allows us to split single sentence fragments into increasingly smaller parts (e.g., causes can be split into variable and condition, which in turn can be decomposed into further more granular text fragments). In this way, we enable a fine-grained causality extraction.
 
\subsection{How do RNTN work?}\label{RNTNalgo}
\begin{figure}
        \centering
         \includegraphics[trim=1cm 5cm 1cm 3cm, width=\linewidth]{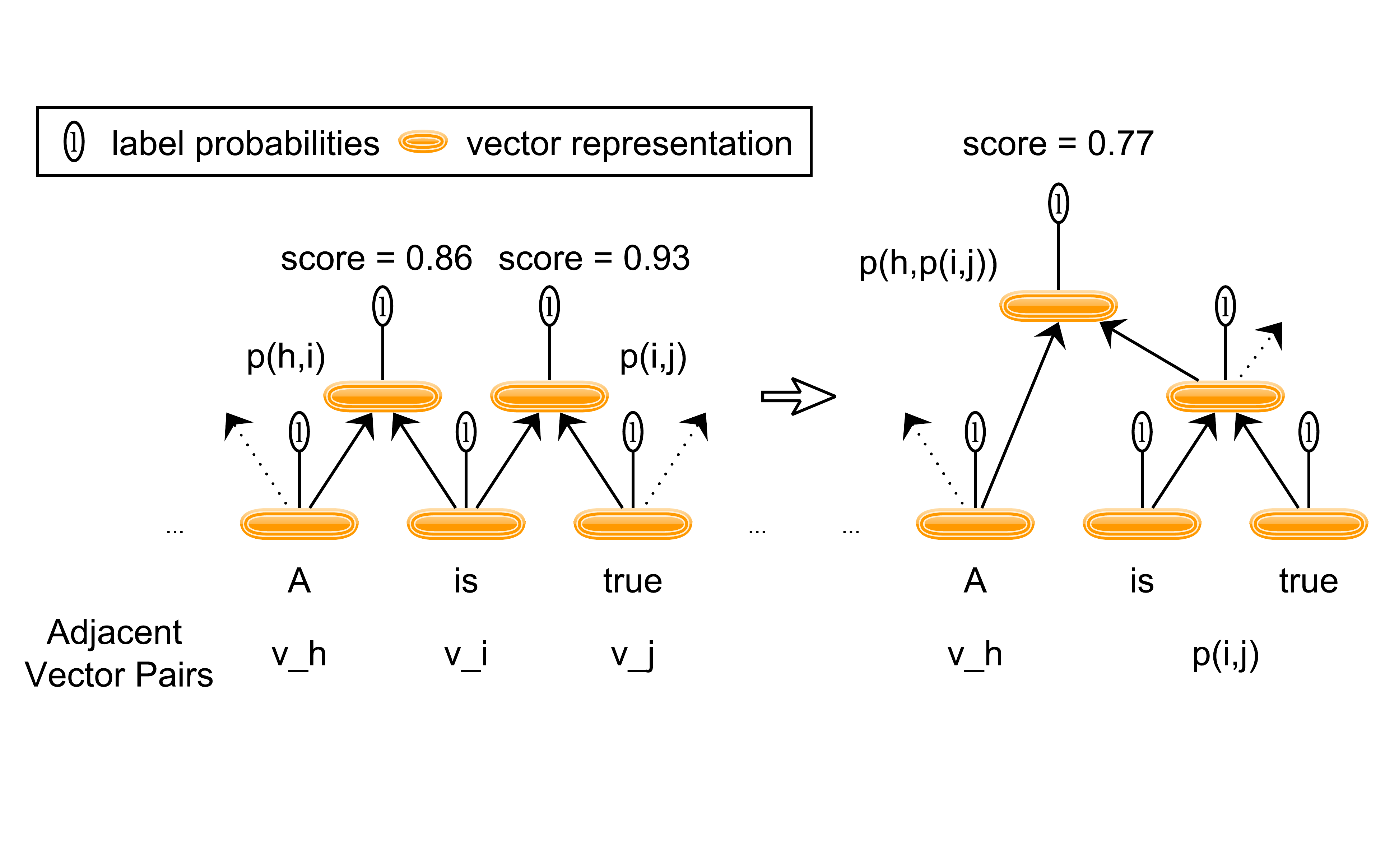}
    \caption{Bottom-up approach of a RNTN for the prediction of a binary parse tree.}\label{functionRNTN}
    \vspace{-.5cm}
    \end{figure}

A RNTN is a special type of neural network and has been invented by Socher et al.~\cite{socher11}. In the following, we explain its characteristics with respect to forward and backward propagation. Let us define a sentence as a sequence of words $(w_i,...,w_n)$, each represented by a corresponding $d$-dimensional vector $v_i$. The concept of a RNTN is to identify related vectors and merge them into pre-defined segments. We train the RNTN, for example, to demarcate vectors that describe causes or effects from non-causal vectors. To illustrate the composition of a sentence, a RNTN builds up a binary tree of segments in a bottom-up fashion (see Fig.~\ref{functionRNTN}).

\textbf{Forward Pass} At the beginning of each recursion, the RNTN determines a set of all adjacent vector pairs $A = \{[v_i,v_j] | v_i \text{ and } v_j \text{ are adjacent}\}$. For each adjacent pair, the RNTN computes: a parent representation $p_{(i,j)}$ and its label probabilities $l$. The parent representation is calculated by merging $v_i$ and $v_j$ according to the following equation:

\small
\begin{equation}
     p_{(i,j)}=f \Bigg( \underbrace{\begin{bmatrix}
         v_i \\
         v_j \\
        \end{bmatrix}^{T} V^{[1:d]} \begin{bmatrix}
         v_i \\
         v_j \\
        \end{bmatrix}}_{\text{Enhancement in RNTN}} + \underbrace{W \begin{bmatrix}
         v_i \\
         v_j \\
        \end{bmatrix}}_{\text{Vanilla RNN}} \Bigg)
  \end{equation}
  \normalsize

A RNTN stores its trainable parameters in a weight matrix $W \in \mathbb{R}^{d \times 2d}$ and a tensor $V \in \mathbb{R}^{2d \times 2d \times d}$. In vanilla Recursive Neural Models (RNN)~\cite{socher11}, $p_i$ is computed only using $W$. The concatenated $v_i$ and $v_j$ are simply multiplied with $W$ and then given into an activation function $f$. This has the drawback that the input vectors only implicitly interact through the nonlinearity activation function. Consequently, the meanings of the words do not actually relate to each other and the influence of a child node on a parent node can not be adequately captured. However, this is especially important for the extraction of causal relations. For example, the model should learn that \textit{if} usually indicates a cause segment, while \textit{then} denotes an effect segment. To overcome this issue, Socher et al.~\cite{socher13} introduced a three-dimensional tensor $V$ into the model. Firstly, $V$ allows more interactions between the vectors through the more direct multiplicative relation. Secondly, $V$ adds another fixed number of parameters to the model, which allows the RNTN to gather even more information about the composition of a sentence. Each slice of the tensor is capturing a specific type of composition making the RNTN capable of understanding the structure of a sentence~\cite{socher13}.  Each $p_{(i,j)}$ is then given to a softmax classifier to compute its label probabilities $l$. In other words, the RNTN calculates the probability for $p_{(i,j)}$ representing a certain segment. For classification into three segments (e.g., cause, effect, non-causal), we compute the softmax score as follows:

\small
\begin{equation}
    l = softmax(C p_{(i,j)})
\end{equation}
\normalsize
$C \in \mathbb{R}^{3 \times d}$ is the classification matrix. After computing the probabilities for all adjacent pairs, the RNTN selects the pair which received the highest softmax score and updates $A$ by removing $v_i$ and $v_j$ and adding $p_{(i,j)}$ (see Fig.~\ref{functionRNTN}). This process is repeated until all adjacent pairs are merged and only one vector is left in $A$. This vector represents the whole sentence and corresponds to the root of the constructed tree-structure.  

\textbf{Backward Pass}
We want to maximize the probability that the RNTN correctly predicts the causes and effects in a sentence. For this purpose, we minimize the cross-entropy error between the predicted and target labels at each node. The backpropagation of a RNTN is slightly more complex than in conventional neural models, because the errors have to be routed through the tree structure (starting from the root). \textit{Backgropagation through structure}~\cite{goller96} is characterized by the following three properties: Firstly, the full derivative for $V$ and $W$ is the sum of the derivatives at each of the nodes. Secondly, the derivatives are split at each node and are sent down both branches of the tree to the next level below. Thirdly, for each node, the error message is the sum of the error propagated from the parent and the error from the node itself. Due to space limitations, we do not specify the exact calculations as we do in the case of the forward pass. Instead, we refer to the paper by Goller and Küchler~\cite{goller96}, which explains \textit{backpropagation through structure} in detail.

\section{Causality Treebank}
To the best of our knowledge, we are the first to utilize a RNTN for RE purposes. Accordingly, there is no labeled corpus of requirements available in the RE community that could be used to train a RNTN. This section describes how we created a suitable training corpus. 

\subsection{Data Collection}
Since we train the RNTN to extract causal relations, we need requirements that follow a causal pattern. Hence, we searched for causal requirements in the data set of 212,186 sentences published by Fischbach et al.~\cite{fischbachRENEXT}. Since a manual analysis of all sentences is not practicable, we searched for specific cue phrases that usually indicate causality~\cite{fischbachREFSQ}. We focused on the cue phrases: \textit{if}, \textit{when}, \textit{in order to}, \textit{due to}, \textit{because}, \textit{since} and \textit{in (the) case of}. We randomly searched the data set for sentences containing these cue phrases, and discussed in the research group whether a sentence (1) represents a functional requirement and (2) exhibits causality. We continued the search until we reached a reasonable number of sentences for the training of the RNTN. As a result, the filtered data set consists of 1,571 causal requirements. 

\subsection{Annotation Process}\label{annotation_process}
\textbf{Labeling}  We need to indicate specific segments in the requirements, so that the RNTN can learn to identify the structure of a causal relation. Since we aim to extract causal relations in fine-grained form, we annotated the 1,571 causal sentences with 27 different segments. To minimize the annotation effort, we analyzed which segments actually need to be annotated manually and which segments can be labeled automatically. We found that 12 of the 27 segments can be assigned automatically. For example, \autour{Word} and \autour{Punct} segments can be rule-based annotated, since they only occur at the lowest level of the tree. 

\textbf{Manually assigned labels:}
\begin{compactenum}
\item[1] \autour{Variable} This label indicates noun phrases in a sentence.
\item[2] \autour{Condition} This label indicates the verb phrases that belong to a variable.
\item[3] \autour{Negation} This label indicates negations (e.g. negated conditions).
\item[4] \autour{Statement} This label indicates combinations of variable and condition segments.
\item[5] \autour{Non-causal} This label indicates segments that are not part of the causal relation.
\item[6] \autour{Key-C} This label highlights certain cue phrases that indicate causality.
\item[7] \autour{Cause} This label indicates a cause segment.
\item[8] \autour{Effect} This label indicates an effect segment.
\item[9] \autour{Cause Effect Relation} This label captures all segments that belong to the causal relation. 
\item[10] \autour{And} This label is used to annotate conjunctions.
\item[11] \autour{Or}  This label is used to annotate disjunctions.
\item[12] \autour{Root Sentence} This label always represents the root node of a sentence.
\item[13] \autour{Key-NC} This label highlights certain cue phrases that indicate non-causal segments in a sentence.
\item[14] \autour{Insertion} This label is used to annotate any kinds of insertions in a sentence (e.g., bracket expressions that are not essential for the interpretation of a sentence but provide additional information).
\item[15] \autour{Sentence} This label is used to connect, e.g., non-causal and causal segments of a sentence. Contrary to the \autour{Root Sentence} segment, it does not contain the ending punctuation mark of the sentence.
\end{compactenum}

\textbf{Automatically assigned labels:}
\begin{compactenum}
\item[1-9]  \autour{Separated...} \{Cause | Statement | And | Cause Effect Relation | Negation | Non-causal | Or | Effect | Variable\}: This label is used to highlight self-contained text fragments that are syntactically separated from other fragments. The label is needed, for example, when a \autour{Statement} segment will be merged with a comma token. The comma turns the segment into a \autour{Separated Statement}.
\item[10] \autour{Word} This label is distributed at the bottom level of the tree and assigned to the individual words of a sentence.
\item[11] \autour{Punct} This label is used to indicate punctuation marks.
\item[12] \autour{Symbol} This label is used to mark special symbols.
\end{compactenum}

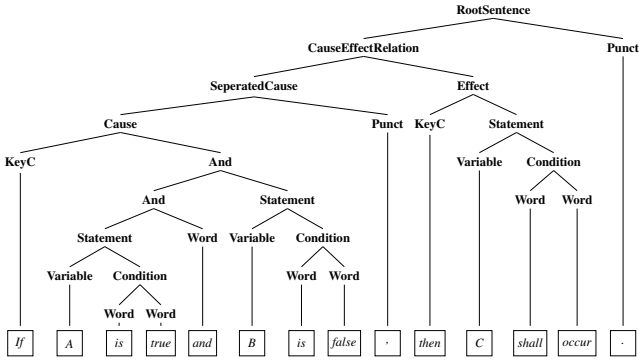
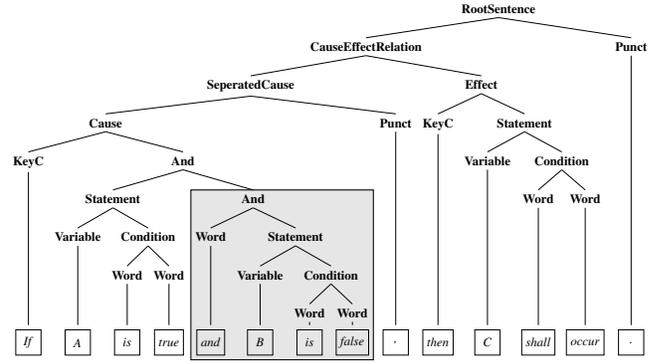
\begin{figure*}
    \centering
    \begin{subfigure}[t]{0.5\textwidth}
        \centering
        \begin{tikzpicture}[scale=.48]
        \tikzset{frontier/.style={distance from root=270pt}}
         \Tree[.\textbf{RootSentence} [.\textbf{CauseEffectRelation} [.\textbf{SeperatedCause} [.\textbf{Cause} [.\textbf{KeyC} \addsquare{\textit{If}} ] [.\textbf{And} [.\textbf{And} [.\textbf{Statement} [.\textbf{Variable} \addsquare{\textit{A}} ] [.\textbf{Condition}  [.\textbf{Word} \addsquare{\textit{is}} ] [.\textbf{Word} \addsquare{\textit{true}} ] ]] [.\textbf{Word} \addsquare{\textit{and}} ]]
                           [.\textbf{Statement} [.\textbf{Variable} \addsquare{\textit{B}} ] [.\textbf{Condition} [.\textbf{Word} \addsquare{\textit{is}} ] [.\textbf{Word} \addsquare{\textit{false}} ]  ] ]]]
                           [.\textbf{Punct} \addsquare{\textit{,}} ]] [.\textbf{Effect} [.\textbf{KeyC} \addsquare{\textit{then}} ] [.\textbf{Statement} [.\textbf{Variable} \addsquare{\textit{C}} ] [.\textbf{Condition}  [.\textbf{Word} \addsquare{\textit{shall}} ] [.\textbf{Word} \addsquare{\textit{occur}} ]   ]  ]]
                ]
          [.\textbf{Punct} \addsquare{\textit{.}} ]]
          \end{tikzpicture}
    \caption{Left Branching.}\label{leftbranching}
    \end{subfigure}%
    \begin{subfigure}[t]{0.5\textwidth}
        \centering
        \begin{tikzpicture}[scale=.48]
        \tikzset{frontier/.style={distance from root=270pt}}
         \Tree[.\textbf{RootSentence} [.\textbf{CauseEffectRelation} [.\textbf{SeperatedCause} [.\textbf{Cause} [.\textbf{KeyC} \addsquare{\textit{If}} ] [.\textbf{And}  [.\textbf{Statement} [.\textbf{Variable} \addsquare{\textit{A}} ] [.\textbf{Condition}  [.\textbf{Word} \addsquare{\textit{is}} ] [.\textbf{Word} \addsquare{\textit{true}} ] ]]
        [.\node(and1){\textbf{And}};  [.\node(and2){\textbf{Word}}; \node(and3){\addsquare{\textit{and}}}; ]               [.\node(statement){\textbf{Statement}}; [.\node(variable){\textbf{Variable}}; \addsquare{\textit{B}} ] [.\node(condition){\textbf{Condition}}; [.\textbf{Word} \addsquare{\textit{is}} ] [.\textbf{Word} \node(word){\addsquare{\textit{false}}}; ]  ] ] ] ]]
                           [.\textbf{Punct} \addsquare{\textit{,}} ]] [.\textbf{Effect} [.\textbf{KeyC} \addsquare{\textit{then}} ] [.\textbf{Statement} [.\textbf{Variable} \addsquare{\textit{C}} ] [.\textbf{Condition} [.\textbf{Word} \addsquare{\textit{shall}} ]  [.\textbf{Word} \addsquare{\textit{occur}} ]  ]  ]]
                ]
          [.\textbf{Punct} \addsquare{\textit{.}} ]]
          \begin{scope}[on background layer]
          \node[draw,fill=gray!20,inner sep=0pt,fit=(and1)(and2)(statement)(variable)(condition)(and3)(word)]{};
          \end{scope}
          \end{tikzpicture}
    \caption{Right Branching.}\label{rightbranching}
    \end{subfigure}
    \caption{Correct binary parse of the requirement: \textit{If A is true and B is false, then C shall occur.} Segments are highlighted in bold. Tokens are indicated by frames. Left: Parse tree after applying left-branching. Right: Parse tree after applying right-branching. Grey highlighting demonstrates the difference between both branching methods.}\label{branchingWays}
   \vspace{-.5cm}
\end{figure*}

\textbf{Annotation Procedure}
As described in Section~\ref{RNTNalgo}, a RNTN builds the tree-structure in a bottom-up fashion. In order to reflect the composition of a sentence, it tries to identify tokens that belong contextually together and merges them into a segment. When adding new tokens to a segment, it decides what information is added by the new token and whether the label of the segment needs to be adjusted or not. We considered this approach during the annotation process and wrote an annotation guideline specifying five steps according to which the sentences should be labeled. We involved four annotators and conducted a workshop where we discussed several examples. Since the quality of the annotations is fundamental for the performance of our final model, we describe the applied annotation procedure in detail. We use \keys{REQ 1} as our sentence to be annotated (see Fig.~\ref{branchingWays}). 

\textit{Step 1: Identify words, punctuation marks, and special symbols.}
The first annotation level is trivial. Each individual word is assigned the label \autour{Word}, while the punctuation marks are annotated with \autour{Punct}. The sentence does not contain any special symbols.

\textit{Step 2: Identify variables and conditions. Also, examine for negations.}
On the second level, we distinguish between variables and conditions. In the present case, \textit{A}, \textit{B} and \textit{C} can be marked as a \autour{Variable} segment, while the verb phrases are labeled as \autour{Condition} segments. None of the conditions is negated.

\textit{Step 3: Identify statements and understand combinatorics.}
Most challenging is the annotation of the third level, which can be illustrated by the adjacent pair $[\text{A},\text{is true}]$. According to the second annotation level, \textit{A} represents a \autour{Variable} segment, while \textit{is true} is part of a \autour{Condition} segment. Combined, the two segments form the expression \textit{A is true}. As readers of the sentence, we know due to the preceding \textit{if} phrase that this expression represents a cause and should be annotated with a corresponding label. However, since the RNTN builds the tree in a bottom-up fashion, it is unable to take the cue phrase into account when merging $[\text{A},\text{is true}]$. The content of both segments does not allow any conclusion about causation. In fact, both segments could also be part of an \autour{Effect} segment or even part of a \autour{Non-causal} segment. However, the combination of a noun phrase and verb phrase allows us to infer a \autour{Statement} segment. Similarly, the adjacent pair $[\text{B},\text{is false}]$ can be also annotated as a \autour{Statement}. To cover the combinatorics between both statements, they must be connected by a conjunction. However, this is not directly possible, because both segments are not adjacent and interrupted by an \textit{and} token. This results in two options: We can either merge \textit{and} with the left neighbor (left branching, see Fig.~\ref{leftbranching}) or with the right neighbor (right branching, see Fig.~\ref{rightbranching}). In our paper, we experimented with both branching methods and implemented them in our exporter (see lines 24 - 38 in Algorithm~\ref{algoExporter}). In the case of \keys{REQ 1}, we assume that right branching is used. Consequently, we label the expression \textit{and B is false} as an \autour{And} segment, because the added \textit{and} token turns the statement into a conjunctive statement (see grey highlighting in Fig.~\ref{rightbranching}).

\textit{Step 4: Identify causes and effects.}
The RNTN can only recognize that the conjunctive statement represents a cause by merging the adjacent pair $[\text{If},\text{A is true and B is false}]$. The cue phrase \textit{if} provides valuable information to the model, which consequently changes the label of the segment from \autour{Statement} to \autour{Cause}. The same applies if you combine $[\text{then},\text{C shall occur}]$. The cue phrase \textit{then} indicates that this segment represents an \autour{Effect}.

\textit{Step 5: Connect cause-effect-pairs.}
In the last step, the related cause and effect pairs are joined into a \autour{Cause Effect Relation} segment. \keys{REQ 1} contains only a single cause effect relation. In some cases, however, a sentence may contain several causal relations, so that several \autour{Cause Effect Relation} segments must be annotated (see [S4] in Tab.~\ref{bratannotations}).

\begin{table*} 
    \caption{Examples of manual annotations and their corresponding binary structured output. Segment names are represented by numbers in the binary annotations (e.g., a word segment is indicated by 23). The exact mapping of segment name to corresponding number is available in our github repository.}
\centering 
\begin{tabular}{ >{\raggedright} m{0.5cm} m{9cm} m{7.5cm}  }      
\toprule                                   
 & Manual Annotations in \textit{brat} & Binary Structured Annotations Generated by Our Exporter \\  
\midrule\addlinespace[1.5ex]
[S1] &
         \includegraphics[width=0.5\textwidth]{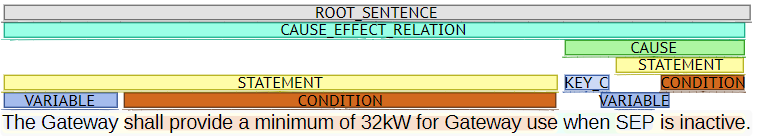} & (1 (13 (10 (9 (23 The) (23 Gateway)) (8 (8 (8 (8 (8 (8 (8 (8 (23 shall) (23 provide)) (23 a)) (23 minimum)) (23 of)) (23 32kW)) (23 for)) (23 Gateway)) (23 use))) (11 (6 when) (10 (9 SEP) (8 (23 is) (23 inactive)))))(3 .))
        \\
\midrule\addlinespace[1.5ex]
[S2] &
         \includegraphics[width=0.5\textwidth]{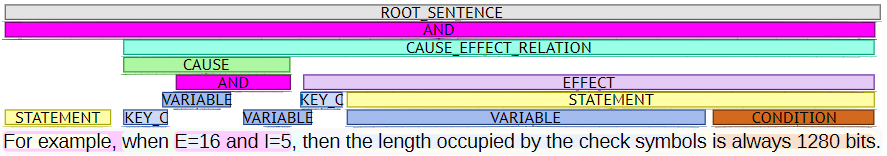} & (1 (20 (20 (17 (23 For) (23 example))(3 ,)) (13 (14 (11 (6 when) (4 (4 (9 E=16) (23 and)) (9 I=5)))(3 ,)) (12 (6 then) (10 (9 (9 (9 (9 (9 (9 (23 the) (23 length)) (23 occupied)) (23 by)) (23 the)) (23 check)) (23 symbols)) (8 (8 (8 (23 is) (23 always)) (23 1280)) (23 bits))))))(3 .))
        \\
        \midrule\addlinespace[1.5ex]
        [S3] &
         \includegraphics[width=0.5\textwidth]{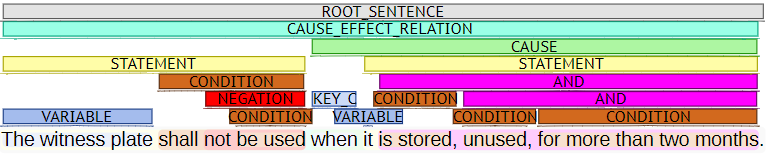} & (1 (13 (10 (9 (9 (23 The) (23 witness)) (23 plate)) (16 (16 (23 shall) (23 not)) (8 (23 be) (23 used)))) (11 (6 when) (10 (9 it) (4 (4 (8 (23 is) (23 stored))(3 ,)) (4 (4 (8 unused)(3 ,)) (8 (8 (8 (8 (23 for) (23 more)) (23 than)) (23 two)) (23 months)))))))(3 .))
        \\
         \midrule\addlinespace[1.5ex]
        [S4] &
         \includegraphics[width=0.5\textwidth]{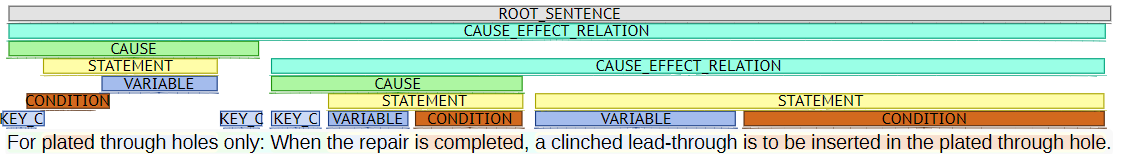} & (1 (20 (20 (17 (17 (17 (17 (23 For) (23 plated)) (23 through)) (23 holes)) (23 only))(2 :)) (13 (14 (11 (6 When) (10 (9 (23 the) (23 repair)) (8 (23 is) (23 completed))))(3 ,)) (10 (9 (9 (23 a) (23 clinched)) (23 lead-through)) (8 (8 (8 (8 (8 (8 (8 (8 (23 is) (23 to)) (23 be)) (23 inserted)) (23 in)) (23 the)) (23 plated)) (23 through)) (23 hole)))))(3 .))
        \\
\bottomrule
    \end{tabular}
    \label{bratannotations}
   \vspace{-.5cm}
\end{table*}

\textbf{Annotation Tool \& Tree Exporter}
A RNTN requires a strongly structured training input. More specifically, it needs to be trained on binary trees. This renders the annotation of individual segments laborious and error-prone. To support the annotators, we used the web-based \textit{brat}\footnote{Since \textit{brat} was not originally designed for annotating multiple layers, we slightly modified the platform. We share the customized code in our github repository.} annotation platform~\cite{brat12}. Instead of requiring the annotators to assign separate segment labels for each adjacent token pair, we allow the annotation of segments that span multiple tokens (see Tab.~\ref{bratannotations}). Consequently, the manual annotations do not need to follow a binary structure, allowing the annotation process to be more efficient. In order to subsequently convert the annotations into a format usable for training the RNTN, we implement a post-processing step. Specifically, we rebuild the binary structure for each annotated sentence. To this end, we implement an exporter, which transforms the annotations into a binary structured output. This can illustrated by [S2] in Tab.~\ref{bratannotations}. The four tokens \textit{is always 1290 bits} are marked as a single \autour{Condition} segment. However, a RNTN would expect three condition labels in this case, i.e. for the adjacent pairs $[\text{is},\text{always}]$, $[\text{is always},\text{1290}]$ and $[\text{is always 1290},\text{bits}]$ if we apply left-branching. Our exporter sets these labels automatically and marks segments with brackets. The segment labels are represented by numbers to minimize the length of the binary annotations. Thus, the exporter creates the following binary annotation for the expression \textit{is always 1290 bits}:

\begin{center}
(8 (8 (8 (23 is) (23 always)) (23 1280))(23 bits))
\end{center}

\autour{Condition} segments are tagged as 8, while \autour{Word} segments are marked as 23. The exact functionality of our exporter is defined by Algorithm~\ref{algoExporter}. We share the code of the exporter in our github repository.\textsuperscript{\ref{note1}}

\subsection{Annotation Validity}
To verify the reliability of the manual annotations, we calculated the inter-annotator agreement. For this purpose, we distributed the 1,571 causal sentences among the four annotators, ensuring that 314 sentences are labeled by two annotators (overlapping quote of $\approx$20\%). Similar to other studies~\cite{kolditz19} that also utilize \textit{brat} to annotate text segments, we calculate the pair-wise averaged F1 score~\cite{hripcsak05} based on the overlapping sentences. Specifically, we treat one rater as the subject and the other rater's answers as if they were a gold standard. This allows us to calculate the Precision and Recall values for their annotations. We then determine the F1 score as the harmonic mean ($\beta = 1$) of Recall and Precision:

\small
\begin{equation}
F = \frac{(1+\beta^2)\cdot Recall \cdot Precision}{(\beta^2 \cdot Precision) + Recall}
\end{equation}
\normalsize

We use this formula to calculate the F1 score pairwise between all raters.  Subsequently, we take the average of F1 scores among all pairs of raters in order to quantify the agreement of our raters: The higher the average F1 score, the more the raters agree with each other. For most of our manually assigned labels, we obtained an inter-annotator agreement of at least 0.83. The lowest agreement was achieved for \autour{Condition} and \autour{And} segments (0.73). The annotators did not always agree on how granular some expressions should be labeled (e.g., are there multiple conditions specified that need to be labeled as separate segments or can they be interpreted as one single segment?). The highest agreement was measured for the assignment of \autour{Statement} segments (0.89). Based on the achieved inter-annotator agreement values, we assess our labeled data set as reliable and suitable for the implementation of our causality extraction approach.

\subsection{Data Analysis}
\begin{figure*}
\centering
\fbox{\includegraphics[width=\textwidth]{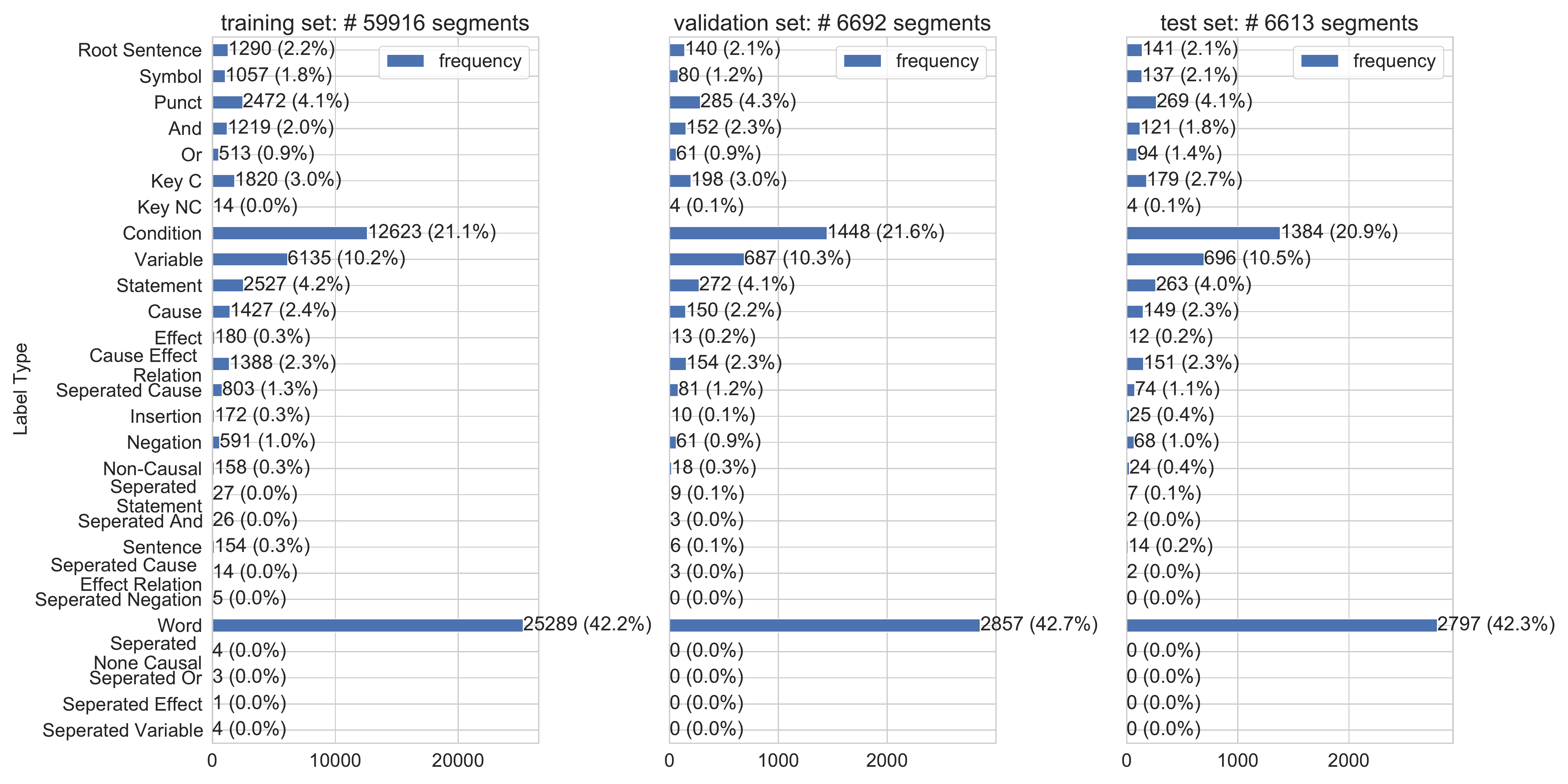}}
\caption{Overview of the segment distribution in our training, validation and testing data sets.}\label{distribution}
\vspace{-.5cm}
\end{figure*}

Our final binary tree structured data set contains a total of 73,221 segments. Fig.~\ref{distribution} provides an overview of the distribution of the segments across the individual labels. The distribution of segments is strongly unbalanced, since some segments (e.g., \autour{Cause Effect Relation}) only occur on the upper levels of binary trees. Most of the segments represent \autour{Words}, \autour{Conditions} and \autour{Variables}, because these labels are already assigned at the lower levels of the tree when multiple smaller text fragments are merged. A \autour{Root Sentence} segment, on the other hand, occurs only once in a sentence. Hence, we find only 1,570 segments with this label in our data set. As shown in Fig.~\ref{distribution}, there are significantly more \autour{Cause} segments than \autour{Effect} segments. One would expect that each cause effect relation contains at least one effect. Consequently, the number of effect segments should be at least equal to the number of cause effect relations. In the formulation of conditional statements, however, effects are rarely explicitly introduced by cue phrases (e.g., \textit{then}). In general, causes are introduced by cue phrases while the effects are implicitly expressed by the semantics of the sentence. This often results in the combination of several \autour{Cause} segments and \autour{Statement} segments, which implicitly express the effect (see S1, S3 and S4 in Tab.~\ref{bratannotations}). Our first experiments have shown that it is important to distinguish between the explicit and implicit form of effects during the annotation process, because otherwise the RNTN gets confused while learning the tree structures in bottom-up fashion. Thus, we annotate \autour{Statement} segments only as \autour{Effect} segments if explicitly indicated by a cue phrase (see Step 4 in Section~\ref{annotation_process}). 

\begin{algorithm}
 \caption{Brat Annotation Export Algorithm}\label{algoExporter}
        \BlankLine
     \KwData{Brat annotation file}
     \KwResult{Binary tree structured sentence}
     \BlankLine
     initialize dataframes\;
     \While{annotations not empty}{
            \ForEach{sentence}{
              convert annotations to dataframe containing begin and end positions of labels\;
              append additional WORDCOUNT column to the data frame capturing the number of individual words in a label\;
              create label containing final clause\;
              add label to dataframe\;
              \eIf{label spanning from beginning of sentence to predecessor of final clause exists}{
                continue\;
              }{
              create label spanning from beginning to predecessor of final clause\;
              add label to dataframe\;
              }
              \ForEach{word in sentence}{
                    \eIf{word has single label}{
                        add label to dataframe\;
                    }{
                        create label WORD spanning from begin to end of word\;
                        add label to dataframe\;
                    }
                }
                \ForEach{label in dataframe}{
                  \If{label has more than two childs}{
                        \eIf{label contains separator}{
                          perform separator merge\;
                        }{
                          \If{leftBranching}{
                              \For{child = 1 to childsOfLabel}{
                                  \eIf{child+1 == childsOfLabel}{
                                    break\;
                                  }{
                                  create label from child to child+1\;
                                  add label to dataframe\;
                                  }
                              }
                          }
                          \If{rightBranching}{
                                \For{child = childsOfLabel to 1}{
                                    \eIf{child-1 == childsOfLabel}{
                                        break\;
                                    }{
                                    create label from child to child-1\;
                                    add label to dataframe\;
                                    }
                                }
                            }
                        }
                    }
                }
            }
      }\end{algorithm}
   
\section{Experiments and Evaluation}
This section presents the training and evaluation of the RNTN based on the \emph{Causality Treebank}. To determine the optimal configuration of the RNTN, we perform the following steps: First, we tune its hyperparameters (see Section~\ref{hyperparameter}). Second, we run two experiments and investigate whether we can improve the performance of the RNTN by using e.g. word vectors enriched with syntactic information (see Section~\ref{experimentsSetup} and Section~\ref{resultsExperiments}). We then apply the best RNTN model to our test set and analyze how well our model can predict the individual segments (see Section~\ref{offlineTesting}). 

\subsection{Evaluation Procedure}
We follow the idea of Cross Validation and divide the data set (1,571 sentences) in a training (1,290), validation (140) and test (141) set. Each segment is equally represented across all three data sets which helps to avoid bias in the prediction (see Fig.~\ref{distribution}). The training set is used to fit the algorithm while the validation set is used to tune its parameters. The test set is utilized for the evaluation of the algorithm based on real world unseen data. We opt for a 10-fold Cross Validation as several studies have shown that a model that has been trained this way demonstrates low bias and variance~\cite{James13}. We use standard metrics, for evaluating our approaches: Accuracy, Precision, Recall, and F1 score. During the training process, we check the validation accuracy periodically in order to keep the model’s checkpoint with the best validation performance.

\subsection{Hyperparameter Tuning}\label{hyperparameter}
We train the RNTN for 90 epochs seeking the optimal hyperparameter configuration. Specifically, we use AdaGrad as optimizer and set the learning rates (lr) to 0.1, 0.01, 0.001, and 0.0001. In addition, we try different mini batch (mb) sizes: 16, 24, 32, and 64. We set epsilon to 1e-08. As described in Section~\ref{RNTNalgo}, each word needs to be represented as a d-dimensional vector. We try different dimension (wvecDim) sizes: 30, 50, and 60. Similarly to Socher et al.~\cite{socher13}, we initialize all word vectors by randomly sampling each value from a uniform distribution: $\mathcal{U}$(-r, r), where r = 0.0001. Consequently, the word vectors are random at the beginning of the training process. However, we consider the word vectors as parameters that are trained jointly with the other parameters of the RNTN. We achieve the best performance with the following configuration: lr = 0.001, mb = 24 and wvecDim = 60. The model yields a training accuracy of 0.931 and a validation accuracy of 0.913 in epoch 87. 

\subsection{Setup of the Experiments}\label{experimentsSetup}
To further improve the performance of the RNTN, we conducted two experiments based on the identified optimal hyperparameter configuration. 

\begin{figure*}
    \centering
    \begin{subfigure}[t]{0.5\textwidth}
        \centering
         \includegraphics[width=\textwidth]{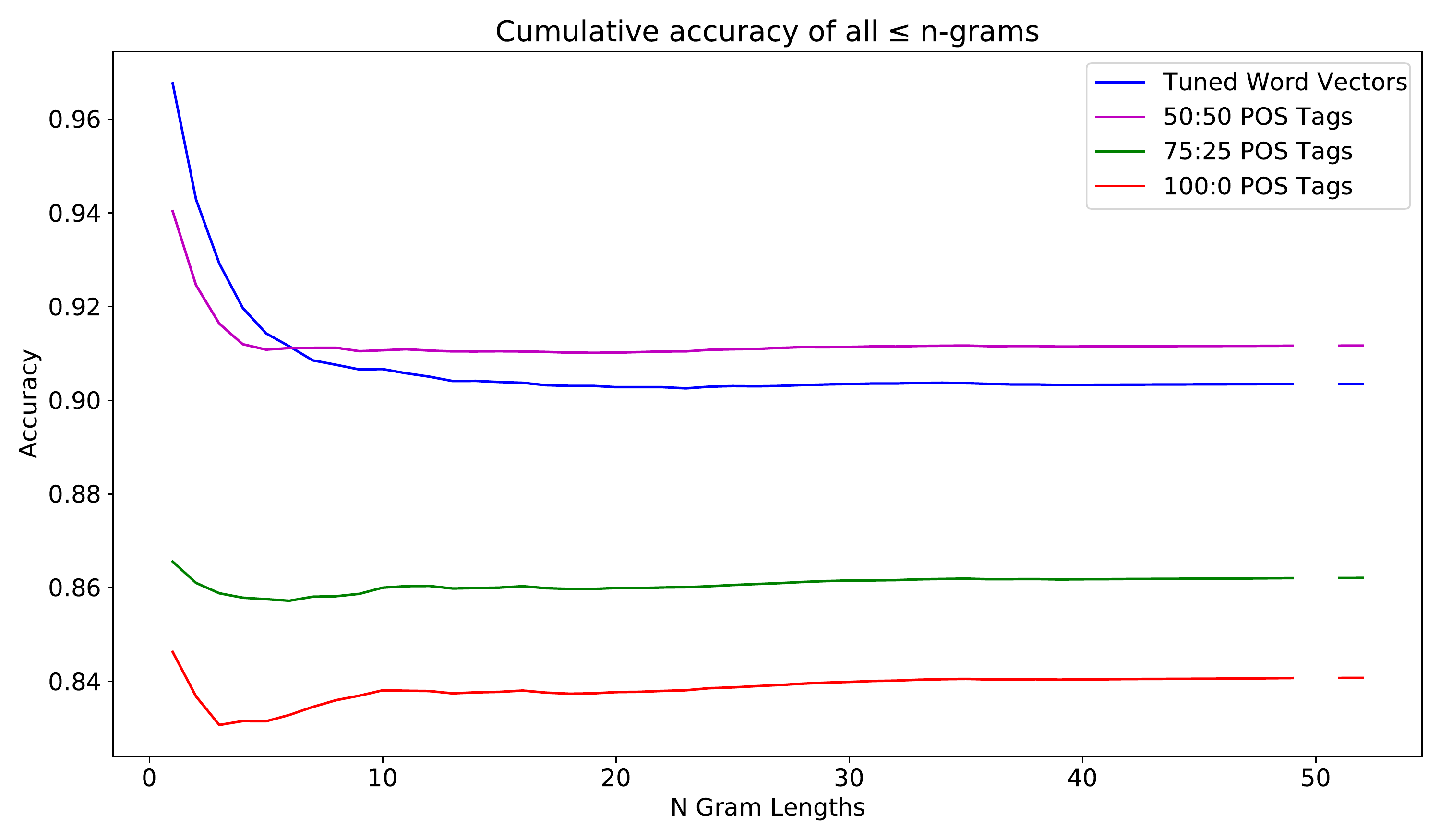}
    \caption{POS Tagging Experiment.}\label{POSPlots}
    \end{subfigure}%
    \begin{subfigure}[t]{0.5\textwidth}
        \centering
         \includegraphics[width=\textwidth]{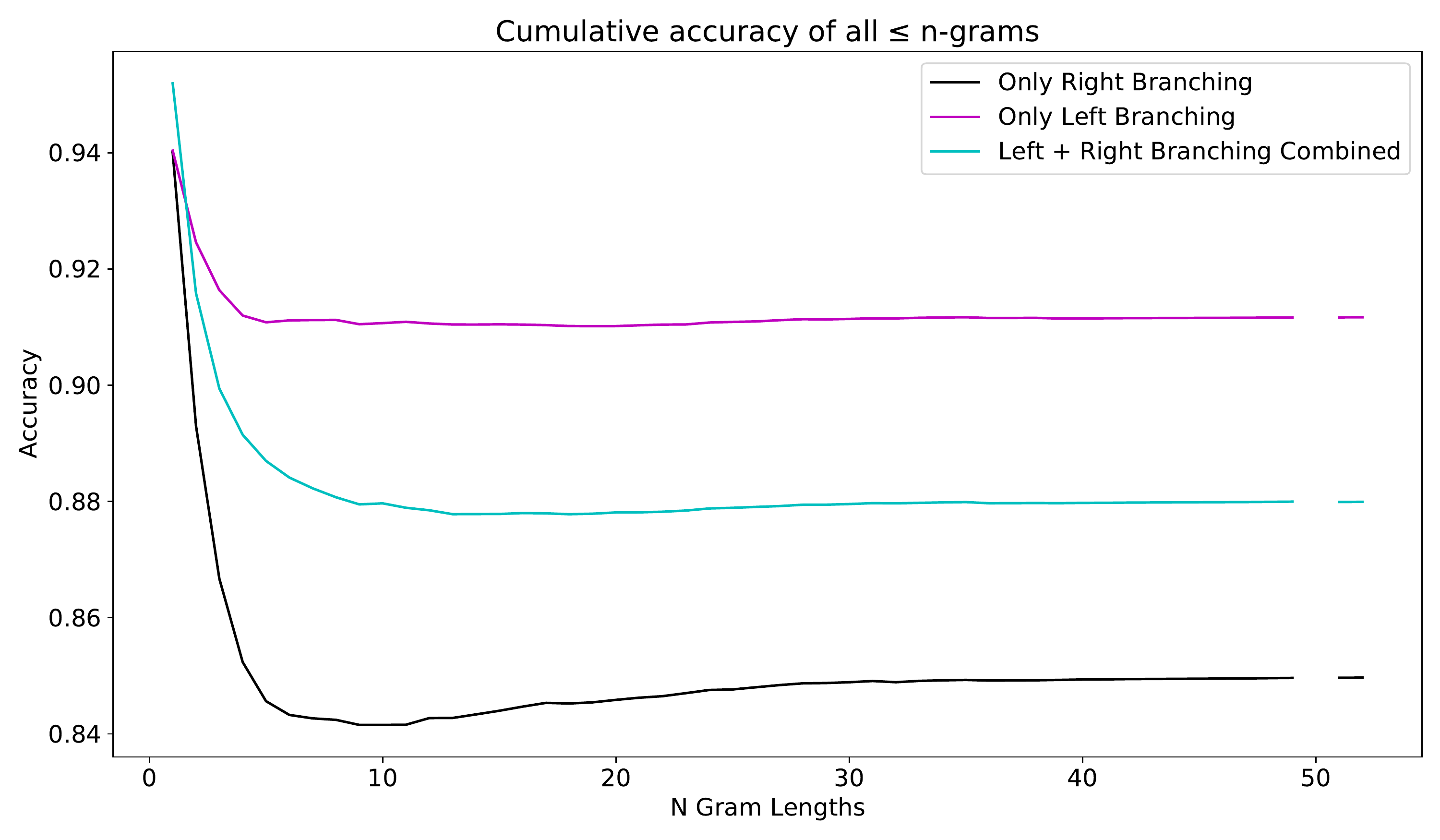}
    \caption{Branching Experiment.}\label{branchingPlots}
    \end{subfigure}
    \caption{Accuracy curves for fine-grained causality extraction at each n-gram lengths. Both plots show the cumulative Accuracy of all $\leq \ $n-grams. Left: Impact of tuned word vectors and syntactically enriched word vectors on performance. Right: Impact of branching methods on performance.}\label{experimentsPlots}
  \vspace{-.5cm}
\end{figure*}

\textbf{POS Tagging Experiment}
Studies have shown that the performance of NLP models can be
improved by providing explicit prior knowledge of syntactic information to the model~\cite{sundararaman2019,fischbachREFSQ}. In this experiment, we investigate whether syntactic information also have a positive impact on the performance of the RNTN. We study two scenarios: First, the word vectors are randomly initialized and used as trainable parameters (as described in Section~\ref{hyperparameter}). Second, the word vectors are not randomly initialized but rather pre-trained and enriched with Part-of-speech (POS) tags. Specifically, we add the corresponding POS tag to each token and create two vector representations: one for the actual token and one for the associated POS tag. We use the \emph{nltk} library~\cite{bird2009natural} to assign the POS tags to the respective tokens and \emph{fastText}~\cite{bojanowski16} to generate the pre-trained vectors. As found during the hyperparameter tuning, the RNTN performs well with a vector dimension of 60. We stick to this dimension size and simply concatenate the pre-trained vector and the POS tag vector to a single representation. To investigate the impact of the added syntactic information on the model performance, we concatenate the two vectors in three different ways. In the first variant, both vectors are equally weighted. The concatenated vector thus contains 30 dimensions representing the POS tag part and 30 dimensions for the pre-trained part (see equation~\ref{eq:50:50}). In the second variant, we weight the syntactic information slightly more, so that the majority of the dimensions constitute the POS tag part (see equation~\ref{eq:75:25}). Specifically, 75\% of the dimensions represent the POS Tag part. In the third variant, the concatenated vector consists only of the POS tag part, i.e. the RNTN predicts only on the basis of POS tags and does not consider the actual token (see equation~\ref{eq:100:0}). To measure the performance of the RNTN, we compare the test accuracy values achieved by using the different word vectors (see Fig.~\ref{POSPlots}). The equations below illustrate the structure of the three word vector types. The POS tag part is highlighted in light blue, while the pre-trained part is marked in dark blue.
\vspace{-0.8cm}
\begin{multicols}{3}
\noindent
\scriptsize
\begin{equation}\label{eq:50:50}
  \left(\begin{array}{c}
    \rowcolor{blue!10}
    x_0  \\
    \rowcolor{blue!10}
    \vdots  \\
   \rowcolor{blue!10}
    x_{29}    \\
    \rowcolor{blue!20}
    x_{30}   \\
    \rowcolor{blue!20}
    \vdots  \\
    \rowcolor{blue!20}
    x_{59} \\
  \end{array}\right)
\end{equation}\break
\begin{equation}\label{eq:75:25}
  \left(\begin{array}{c}
    \rowcolor{blue!10}
    x_0  \\
    \rowcolor{blue!10}
    \vdots  \\
   \rowcolor{blue!10}
    \vdots   \\
    \rowcolor{blue!10}
    x_{44}   \\
    \rowcolor{blue!20}
    \vdots  \\
    \rowcolor{blue!20}
    x_{59} \\
  \end{array}\right)
  \end{equation}\break
  \begin{equation}\label{eq:100:0}
  \left(\begin{array}{c}
    \rowcolor{blue!10}
    x_0  \\
    \rowcolor{blue!10}
    \vdots  \\
   \rowcolor{blue!10}
    \vdots   \\
    \rowcolor{blue!10}
    \vdots   \\
    \rowcolor{blue!10}
    \vdots  \\
    \rowcolor{blue!10}
    x_{59} \\
  \end{array}\right)
\end{equation}
\end{multicols}
\textbf{Branching Experiment}
In this experiment, we study whether the branching method used to generate the binary tree structured data influences the performance of the RNTN. Specifically, we create two data sets with our tree exporter: one data set in which the adjacent pairs are merged by left branching and one data set in which we apply right branching. We also build a third data set that combines the left and right branching data in order to further increase the amount of training instances. We divide each of the three data sets into a training, validation and test set and train the RNTN on these sets. To measure the performance of the RNTN, we compare the respective test accuracy values (see Fig.~\ref{branchingPlots}). We build on the findings of the POS tagging experiment and use the word vectors with a weighting of 50:50, as this results in the best performance of the model.

\subsection{Results of the Experiments}\label{resultsExperiments}

Fig.~\ref{experimentsPlots} reports the results of our experiments. In this section, we interpret the results and select the best performing model as our final causality extractor.

\textbf{Results of the POS Tagging Experiment} 
Irrespective of the selected word vectors, the RNTN achieves a promising result of at least 84\% test accuracy over all n-grams lengths (see  Fig.~\ref{POSPlots}). We achieve the best performance by using word vectors with the POS tag and pre-trained part weighted equally. A comparison of the performance between the three different POS tag weights shows that the higher the proportion of POS tags in the word vector, the lower the test accuracy. In fact, the model achieves a test accuracy of 91.2\% with the 50:50 weighting, 86.2\% with the 75:25 weighting, and the lowest value of 84.1\% with the 100:0 weighting. We hypothesize that only POS tags are not sufficient to comprehend a causal relation since they only reflect the syntax of a sentence, but not its semantics. Hence, the model performs better if it considers both the POS tag and the actual token during prediction. Interestingly, even if the word vectors are randomly initialized and treated as trainable parameters, the RNTN shows a very good test accuracy of 90.4\%. The tuned word vectors outperform the word vectors with a 75:25 weighting by 4.2\% and the word vectors with a 100:0 POS tag weighting by 6.3\%. For segments with a short length of up to 5-grams, the trainable vectors even outperform the 50:50 weighted vectors. With increasing n-gram length, however, the RNTN shows better performance when using the syntactically enriched vectors. Over all test instances, the difference between the two test accuracy values is small (only 0.8\%). Consequently, only a marginal performance gain could be achieved by adding syntactic information to the word vectors. 

\textbf{Results of the Branching Experiment}
Fig.~\ref{branchingPlots} shows the performance of the RNTN depending on the selected branching method. Similar to the previous experiment, the RNTN performs well on all three test sets and achieves at least 85\% test accuracy. Interestingly, the RNTN achieves a better test accuracy when applying left branching rather than right branching (difference of 6.2\%). Combining the left and right branched data sets, the RNTN achieves a test accuracy of 88\%. Hence, our experiment demonstrates that the RNTN is better at building the binary parse tree using left-branching than right-branching.

\begin{tcolorbox}[breakable, enhanced jigsaw]
\textbf{Summary of Experiments:} We found that enriching word vectors with POS tags does not necessarily lead to a significant performance gain. If POS tags account for more than half of the dimensions of the vector, the RNTN performs worse compared to when the vectors are randomly initialized and trained jointly with the model. The RNTN seems to learn left branching better than right branching. 
\end{tcolorbox}

\begin{table}
\caption{Results of the evaluation on the \emph{Causality Treebank}. The segments ``Separated Negation'', ``Separated Non-causal'', ``Separated Or'', ``Separated Effect'' and ``Separated Variable'' are not included in the test set and are therefore not part of the evaluation. F1 scores of at least 0.9 are marked in \textbf{bold}.}\label{tab:testresults}
\centering
\setlength{\extrarowheight}{0pt}
\addtolength{\extrarowheight}{\aboverulesep}
\addtolength{\extrarowheight}{\belowrulesep}
\setlength{\aboverulesep}{0pt}
\setlength{\belowrulesep}{0pt}
\label{tab:testresults}
\resizebox{\columnwidth}{!}{
\begin{tabular}{lrrr} 
\toprule
                                                          & \multicolumn{3}{c}{\textbf{Performance Measures}}   \\
\cmidrule(lr){2-4}\cmidrule(lr){2-4}
\textbf{Label Type}                              & Recall & Precision & F1 - Score         \\
\midrule
\rowcolor[rgb]{0.89,0.89,0.89} Root Sentence   & 1.0       & 1.0           & \textbf{1.0}                             \\
Symbol & 0.2      & 0.58           & 
0.39                                           \\
\rowcolor[rgb]{0.89,0.89,0.89} Punct & 1.0     & 0.93           & 
\textbf{0.97}                                           \\
And                                & 0.82           & 0.65          & 0.74                                                \\
\rowcolor[rgb]{0.89,0.89,0.89} Or                              & 0.79          & 0.94           & 0.87                  \\
Key-C                               & 0.86          & 0.87         & 0.87                                                    \\
\rowcolor[rgb]{0.89,0.89,0.89} Key-NC                                      & 0.0           & 0.0          & 0.0              \\
Condition        & 0.82          & 0.78           & 0.8                                                    \\
\rowcolor[rgb]{0.89,0.89,0.89} Variable         & 0.88           & 0.91           & \textbf{0.9}    \\
Statement                                   & 0.93          & 0.9           & \textbf{0.92}                   \\
\rowcolor[rgb]{0.89,0.89,0.89} Cause                                  & 0.95           & 0.95           & \textbf{0.95}                                                   \\
Effect                                   & 0.83           & 0.83           & 0.83                                                \\
\rowcolor[rgb]{0.89,0.89,0.89} Cause Effect Relation                                   & 0.93           & 0.94           & \textbf{0.94}                                               \\
Separated Cause                                   & 0.98           & 0.96           & \textbf{0.97}                  \\
\rowcolor[rgb]{0.89,0.89,0.89} Insertion                                   & 0.88           & 0.95           & \textbf{0.92}                                                    \\
Negation                                   & 0.94           & 0.92           & \textbf{0.93}                 \\
\rowcolor[rgb]{0.89,0.89,0.89} Non-causal                                   & 0.0           & 0.0           & 0.0                                                     \\
Separated Statement                                   & 0.42          & 1.0           & 0.71                   \\
\rowcolor[rgb]{0.89,0.89,0.89} Separated And                                   & 1.0           & 0.66           & 0.83                                                   \\
Sentence                                   & 0.21          & 0.6           & 0.41                                                    \\
\rowcolor[rgb]{0.89,0.89,0.89} Separated Cause Effect Relation                                  & 0.5           & 0.33           & 0.42                                           \\
Separated Negation                                   & -          & -          & -                                                     \\
\rowcolor[rgb]{0.89,0.89,0.89} Word                                  & 0.99           & 0.93           & \textbf{0.96}                                               \\
Separated Non-Causal                                   & -           & -         & -                                               \\
\rowcolor[rgb]{0.89,0.89,0.89} Separated Or                                   & -           & -        & -                  \\
Separated Effect                                   & -           & -          & -                                                  \\
\rowcolor[rgb]{0.89,0.89,0.89} Separated Variable                                  & -           & -           & -                                                     \\
\bottomrule
\textbf{Mean} & 0.72 & 0.76 &  0.74 \\
\end{tabular}}
 \vspace{-.5cm}
\end{table}

\subsection{Evaluation}\label{offlineTesting}
\begin{figure*}
    \centering
    \begin{subfigure}[t]{0.48\textwidth}
        \centering
        \begin{tikzpicture}[scale=.27]
        \tikzset{frontier/.style={distance from root=570pt}}
         \Tree[.\textbf{RootSentence} [.\textbf{CauseEffectRelation} [.\textbf{SeperatedCause} [.\textbf{Cause} [.\textbf{KeyC} \addsquare{\textit{When}} ] [.\textbf{Statement} [.\textbf{Variable} [.\textbf{Word} \addsquare{\textit{a}} ] [.\textbf{Word} \addsquare{\textit{page}} ] ] [.\textbf{Condition} [.\textbf{Word} \addsquare{\textit{is}} ] [.\textbf{Word} \addsquare{\textit{created}} ] ] ] ] [.\textbf{Punct} \addsquare{\textit{,}} ] ] [.\textbf{Statement} [.\textbf{Variable} [.\textbf{Word} \addsquare{\textit{the}} ] [.\textbf{Word} \addsquare{\textit{user}} ] ] [.\textbf{\textcolor{red}{Condition}} [.\textbf{\textcolor{red}{Condition}} [.\textbf{\textcolor{red}{Condition}} [.\textbf{\textcolor{red}{Condition}} [.\textbf{\textcolor{red}{Condition}} [.\textbf{\textcolor{red}{Condition}} [.\textbf{\textcolor{red}{Condition}} [.\textbf{\textcolor{red}{Condition}} [.\textbf{\textcolor{red}{Condition}} [.\textbf{\textcolor{red}{Condition}} [.\textbf{\textcolor{red}{Condition}} [.\textbf{\textcolor{red}{Condition}} [.\textbf{\textcolor{red}{Condition}} [.\textbf{\textcolor{red}{Condition}} [.\textbf{\textcolor{red}{Condition}} [.\textbf{Word} \addsquare{\textit{in}} ] [.\textbf{Word} \addsquare{\textit{the}} ] ] [.\textbf{Word} \addsquare{\textit{role}} ] ] [.\textbf{Word} \addsquare{\textit{of}} ] ] [.\textbf{Word} \addsquare{\textit{course}} ] ] [.\textbf{Word} \addsquare{\textit{admin}} ] ] [.\textbf{Word} \addsquare{\textit{should}} ] ] [.\textbf{Word} \addsquare{\textit{be}} ] ] [.\textbf{Word} \addsquare{\textit{able}} ] ] [.\textbf{Word} \addsquare{\textit{to}} ] ] [.\textbf{Word} \addsquare{\textit{toggle}} ] ] [.\textbf{Word} \addsquare{\textit{whether}} ] ] [.\textbf{Word} \addsquare{\textit{notifications}} ] ] [.\textbf{Word} \addsquare{\textit{are}} ] ] [.\textbf{Word} \addsquare{\textit{turned}} ] ] [.\textbf{Word} \addsquare{\textit{on}} ] ] ] ] [.\textbf{Punct} \addsquare{\textit{.}} ]]
\end{tikzpicture}
    \caption{Incorrect binary parse of the RNTN. Improperly identified segments are marked in red.}\label{wrongTree}
    \end{subfigure}\hfill
    \begin{subfigure}[t]{0.48\textwidth}
        \centering
        \begin{tikzpicture}[scale=.27]
        \tikzset{frontier/.style={distance from root=400pt}}
         \Tree[.\textbf{RootSentence} [.\textbf{CauseEffectRelation} [.\textbf{SeperatedCause} [.\textbf{Cause} [.\textbf{KeyC} \addsquare{\textit{When}} ] [.\textbf{Statement} [.\textbf{Variable} [.\textbf{Word} \addsquare{\textit{a}} ] [.\textbf{Word} \addsquare{\textit{page}} ] ] [.\textbf{Condition} [.\textbf{Word} \addsquare{\textit{is}} ] [.\textbf{Word} \addsquare{\textit{created}} ] ] ] ] [.\textbf{Punct} \addsquare{\textit{,}} ] ] [.\textbf{\textcolor{Green}{CauseEffectRelation}} [.\textbf{\textcolor{Green}{Cause}} [.\textbf{Variable} [.\textbf{Word} \addsquare{\textit{the}} ] [.\textbf{Word} \addsquare{\textit{user}} ] ] [.\textbf{\textcolor{Green}{Cause}} [.\textbf{\textcolor{Green}{KeyC}} \addsquare{\textit{in}} ] [.\textbf{\textcolor{Green}{Condition}} [.\textbf{\textcolor{Green}{Condition}} [.\textbf{\textcolor{Green}{Condition}} [.\textbf{\textcolor{Green}{Condition}} [.\textbf{Word} \addsquare{\textit{the}} ] [.\textbf{Word} \addsquare{\textit{role}} ] ] [.\textbf{Word} \addsquare{\textit{of}} ] ] [.\textbf{Word} \addsquare{\textit{course}} ] ] [.\textbf{Word} \addsquare{\textit{admin}} ] ] ] ] [.\textbf{\textcolor{Green}{Condition}} [.\textbf{\textcolor{Green}{Condition}} [.\textbf{\textcolor{Green}{Condition}} [.\textbf{\textcolor{Green}{Condition}} [.\textbf{\textcolor{Green}{Condition}} [.\textbf{\textcolor{Green}{Condition}} [.\textbf{\textcolor{Green}{Condition}} [.\textbf{\textcolor{Green}{Condition}} [.\textbf{\textcolor{Green}{Condition}} [.\textbf{Word} \addsquare{\textit{should}} ] [.\textbf{Word} \addsquare{\textit{be}} ] ] [.\textbf{Word} \addsquare{\textit{able}} ] ] [.\textbf{Word} \addsquare{\textit{to}} ] ] [.\textbf{Word} \addsquare{\textit{toggle}} ] ] [.\textbf{Word} \addsquare{\textit{whether}} ] ] [.\textbf{Word} \addsquare{\textit{notifications}} ] ] [.\textbf{Word} \addsquare{\textit{are}} ] ] [.\textbf{Word} \addsquare{\textit{turned}} ] ] [.\textbf{Word} \addsquare{\textit{on}} ] ] ] ] [.\textbf{Punct} \addsquare{\textit{.}} ]]
\end{tikzpicture}
    \caption{Correct binary parse. Segments that the RNTN did not detect are highlighted in green.}\label{correctTree}
    \end{subfigure}
    \caption{Overview of binary parse trees representing the sentence \textit{When a page is created, the user in the role of course admin should be able to toggle whether notifications are turned on.} Left: Prediction of our trained RNTN. Right: Semantically correct binary parse tree.}\label{wrongParsing}
   \vspace{-.5cm}
\end{figure*}
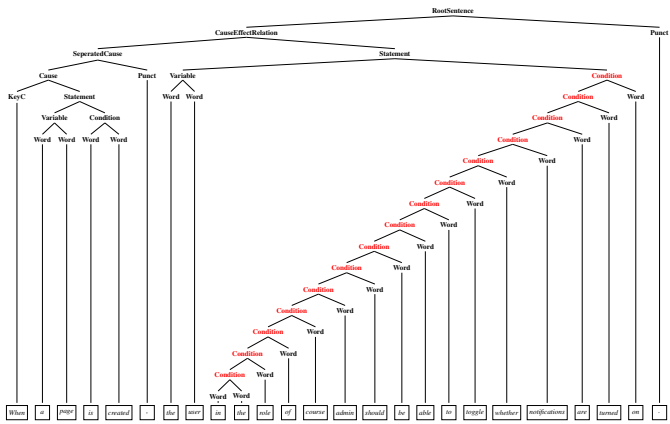
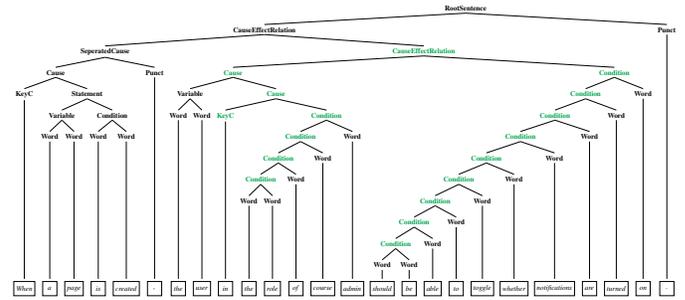

The test accuracy values achieved in our experiments already indicate that the RNTN is able to parse causal relations. However, we are not only interested in the overall test accuracy, but also in the performance with respect to the individual segments. Tab.~\ref{tab:testresults} presents the Recall, Precision and F1 scores per segment. We observe that the RNTN predicts a number of segments reliably. It achieves a F1 score of at least 90\% for the segments \autour{Root Sentence}, \autour{Punct}, \autour{Variable}, \autour{Statement}, \autour{Cause}, \autour{Cause Effect Relation}, \autour{Separated Cause}, \autour{Insertion}, \autour{Negation} and \autour{Word}.
Across all segments, our approach yields a F1 score of 74\%. Not surprisingly, the RNTN is able to predict segments like \autour{Root Sentence}, \autour{Punct} and \autour{Word} almost perfectly, since these segments always occur on the same level of the tree: \autour{Root Sentence} on top and \autour{Punct} and \autour{Word} at the bottom. In addition, \autour{Punct} and \autour{Word} segments always represent 1-grams. Contrary, the RNTN shows a poor performance for the segments \autour{Key NC}, \autour{Non-causal}, \autour{Sentence} and \autour{Separated Cause Effect Relation}. We hypothesize that the poor performance stems from the fact that these segments are highly under-represented in the training and validation set (see Fig.~\ref{distribution}), rendering them difficult for the RNTN to learn. The RNTN seems to detect well which tokens in a NL sentence represent a cause. However, it shows a weaker performance in predicting effects (12\% difference in both Recall and Precision). We hypothesize that this results from the significant under-representation of the explicit form of effects in contrast to its implicit form (see Fig.~\ref{distribution}). The strong performance with respect to the \autour{Cause Effect Relation} segment (F1 score of 94\%) shows that the RNTN acquired the concept of causal relations being a combination of single/multiple \autour{Cause} and \autour{Effect} segments. Interestingly, the RNTN achieves a better F1 score for the prediction of \autour{Or} segments than for \autour{And} segments. Predicting \autour{And} segments seems to be a more difficult task than the prediction of \autour{Or} segments, because the latter usually contain an \textit{or} token, while \autour{And} segments often contain several \autour{Condition} segments which are concatenated without an \textit{and} token (see sentence 3 in Tab.~\ref{bratannotations}). In these cases, the conjunction is implicitly contained in the semantics of the sentence and not by an explicit \textit{and} token, making the prediction challenging. When analyzing the test predictions, we found that the RNTN sometimes fails to distinguish between \autour{Variable} and \autour{Condition} segments, i.e. it assigns tokens that actually belong to a \autour{Variable} segment to a  \autour{Condition} segment and vice versa. Across all test predictions, we observed that the RNTN has a slight bias towards \autour{Condition} segments and tends to construct large separate \autour{Condition} segments. This leads to a significant number of false positives (Precision value of only 78\%) and can even result in a complete false parse tree as indicated by Fig.~\ref{wrongTree}. In this example, the RNTN detects only the outer cause (\textit{a page is created}) and ignores that the sentence contains a second cause: only users with admin rights are allowed to view the notification settings. Rather, the RNTN constructs a large distinct \autour{Condition} segment and merges it with the \autour{Variable} segment $[\text{the user}]$. As a result, the binary tree is assembled incorrectly, because the inner causal relation is not recognized. This example illustrates one of the major limitations of the RNTN. Due to the bottom-up construction of the tree, prediction errors on the lower layers strongly affect the upper layers. Initial experiments revealed that this constitutes a problem especially when we apply the RNTN to words that are not yet part of its training's vocabulary. In such cases, the RNTN is unsure already on the lower layers to which segments the unknown tokens should be assigned and struggles to understand the semantics of the sentence. These errors are propagated to the upper layers meaning that the RNTN builds the binary parse tree based on wrong segments.

\begin{tcolorbox}[breakable, enhanced jigsaw]
\textbf{Summary of Evaluation:} The RNTN predicts most of the segments with high Precision and Recall. It shows a better performance in detecting causes than explicit effects. We see potential for optimization in the detection of conjunctions as well as in the splitting of causes and effects into variable and condition. A threat to the practicability of our approach is that early prediction errors have a major negative impact on the composition of the upper layers. 
\end{tcolorbox}

\section{Conclusions and Outlook}
Causation is a widely used linguistic pattern to describe the expected system behavior in functional requirements (e.g., \textit{If A and B, then C}). Automatically extracting these causal relations supports at least two RE use cases: the automatic derivation of suitable test cases, and the automatic detection of requirements dependencies. However, existing approaches fail to extract causal relations with reasonable performance. Moreover, they extract causal relations only in a coarse-grained form, making them unsuitable for the above-mentioned use cases. We address this research gap and propose a fine-grained causality extractor based on a RNTN. We train the RNTN on our self-annotated data set that consists of 1,571 causal requirements: the \emph{Causality Treebank}. Our data set is the first corpus of fully labeled binary parse trees representing the composition of causal relations in functional requirements. Our evaluation promotes the feasability of our approach. Specifically, our trained RNTN is capable of recovering the composition of a causal relation by detecting 27 different segments (e.g., variables, conditions, causes) in an NL sentence.
Nevertheless, our evaluation also revealed a major limitation of the RNTN, which poses a threat to its applicability in practice. Due to the bottom-up construction of the binary parse tree, prediction errors on the lower layers are propagated to the upper layers, causing partial or complete misinterpretation of the composition. Future work should therefore focus on improving the robustness of the presented approach. Currently, we are working on combining our approach with pre-trained BERT embeddings in order to make the RNTN more robust for the prediction of words that are not yet in its training vocabulary. So far, our approach is limited to the extraction of \textit{explicit} causality. Future work should aim to extend the scope of extraction also to \textit{implicit} causality.

\bibliographystyle{IEEEtran}
\bibliography{references}

\begin{thebibliography}{10}
\providecommand{\url}[1]{#1}
\csname url@samestyle\endcsname
\providecommand{\newblock}{\relax}
\providecommand{\bibinfo}[2]{#2}
\providecommand{\BIBentrySTDinterwordspacing}{\spaceskip=0pt\relax}
\providecommand{\BIBentryALTinterwordstretchfactor}{4}
\providecommand{\BIBentryALTinterwordspacing}{\spaceskip=\fontdimen2\font plus
\BIBentryALTinterwordstretchfactor\fontdimen3\font minus
  \fontdimen4\font\relax}
\providecommand{\BIBforeignlanguage}[2]{{%
\expandafter\ifx\csname l@#1\endcsname\relax
\typeout{** WARNING: IEEEtran.bst: No hyphenation pattern has been}%
\typeout{** loaded for the language `#1'. Using the pattern for}%
\typeout{** the default language instead.}%
\else
\language=\csname l@#1\endcsname
\fi
#2}}
\providecommand{\BIBdecl}{\relax}
\BIBdecl

\bibitem{Mich04}
L.~Mich, M.~Franch, and P.~{Novi Inverardi}, ``Market research for requirements
  analysis using linguistic tools,'' \emph{Requirements Engineering}, 2004.

\bibitem{Kassab14}
M.~Kassab, C.~Neill, and P.~Laplante, ``State of practice in requirements
  engineering: contemporary data,'' \emph{Innovations in Systems and Software
  Engineering}, 2014.

\bibitem{fischbachREFSQ}
J.~Fischbach, J.~Frattini, A.~Spaans, M.~Kummeth, A.~Vogelsang, D.~Mendez, and
  M.~Unterkalmsteiner, ``Automatic detection of causality in requirement
  artifacts: the cira approach,'' in \emph{REFSQ'21}.

\bibitem{fischbachICST}
J.~Fischbach, A.~Vogelsang, D.~Spies, A.~Wehrle, M.~Junker, and
  D.~Freudenstein, ``Specmate: Automated creation of test cases from acceptance
  criteria,'' in \emph{ICST'20}.

\bibitem{fischbachRENEXT}
J.~{Fischbach}, B.~{Hauptmann}, L.~{Konwitschny}, D.~{Spies}, and
  A.~{Vogelsang}, ``Towards causality extraction from requirements,'' in
  \emph{RE'20}.

\bibitem{Garcia97}
D.~Garcia, ``Coatis, an nlp system to locate expressions of actions connected
  by causality links,'' in \emph{EKAW'97}.

\bibitem{khoo98}
C.~S.~G. Khoo, J.~Kornfilt, R.~N. Oddy, and S.~H. Myaeng, ``{Automatic
  Extraction of Cause-Effect Information from Newspaper Text Without
  Knowledge-based Inferencing},'' \emph{Literary and Linguistic Computing},
  1998.

\bibitem{puente}
C.~Puente and J.~A. Olivas, ``Analysis, detection and classification of certain
  conditional sentences in text documents,'' in \emph{IPMU'08}.

\bibitem{Asghar16}
N.~Asghar, ``Automatic extraction of causal relations from natural language
  texts: A comprehensive survey,'' \emph{ArXiv}, vol. abs/1605.07895, 2016.

\bibitem{yang2021survey}
J.~Yang, S.~C. Han, and J.~Poon, ``A survey on extraction of causal relations
  from natural language text,'' \emph{CoRR}, vol. abs/2101.06426, 2021.

\bibitem{Ponti17}
E.~M. Ponti and A.~Korhonen, ``Event-related features in feedforward neural
  networks contribute to identifying causal relations in discourse,'' in
  \emph{EMNLP'17}.

\bibitem{canasai17}
C.~Kruengkrai, K.~Torisawa, C.~Hashimoto, J.~Kloetzer, J.-H. Oh, and M.~Tanaka,
  ``Improving event causality recognition with multiple background knowledge
  sources using multi-column convolutional neural networks,'' in
  \emph{AAAI'17}.

\bibitem{Jin2020}
X.~Jin, X.~Wang, X.~Luo, S.~Huang, and S.~Gu, ``Inter-sentence and implicit
  causality extraction from chinese corpus,'' \emph{Advances in Knowledge
  Discovery and Data Mining}, 2020.

\bibitem{Greene06}
D.~Greene and P.~Cunningham, ``Practical solutions to the problem of diagonal
  dominance in kernel document clustering,'' in \emph{ICML'06}.

\bibitem{ferrari17}
A.~{Ferrari}, G.~O. {Spagnolo}, and S.~{Gnesi}, ``{PURE}: {A} dataset of public
  requirements documents,'' in \emph{RE'17}.

\bibitem{chang05}
D.-S. Chang and K.-S. Choi, ``Causal relation extraction using cue phrase and
  lexical pair probabilities,'' in \emph{IJCNLP'05}, K.-Y. Su, J.~Tsujii, J.-H.
  Lee, and O.~Y. Kwong, Eds.

\bibitem{rink10}
B.~Rink and S.~Harabagiu, ``{UTD}: Classifying semantic relations by combining
  lexical and semantic resources,'' in \emph{SemEval'10}.

\bibitem{dasgupta18}
T.~Dasgupta, R.~Saha, L.~Dey, and A.~Naskar, ``Automatic extraction of causal
  relations from text using linguistically informed deep neural networks,'' in
  \emph{SIGDIAL'18}.

\bibitem{li19}
Z.~Li, Q.~Li, X.~Zou, and J.~Ren, ``Causality extraction based on
  self-attentive {BiLSTM-CRF} with transferred embeddings,'' \emph{CoRR}, vol.
  abs/1904.07629, 2019.

\bibitem{socher13}
R.~Socher, A.~Perelygin, J.~Wu, J.~Chuang, C.~D. Manning, A.~Ng, and C.~Potts,
  ``Recursive deep models for semantic compositionality over a sentiment
  treebank,'' in \emph{EMNLP'13}.

\bibitem{socher11}
R.~Socher, C.~C.-Y. Lin, A.~Y. Ng, and C.~D. Manning, ``Parsing natural scenes
  and natural language with recursive neural networks,'' in \emph{ICML'11}.

\bibitem{goller96}
C.~{Goller} and A.~{Kuchler}, ``Learning task-dependent distributed
  representations by backpropagation through structure,'' in \emph{ICNN'96}.

\bibitem{brat12}
P.~Stenetorp, S.~Pyysalo, G.~Topi{\'c}, T.~Ohta, S.~Ananiadou, and J.~Tsujii,
  ``brat: a web-based tool for {NLP}-assisted text annotation,'' in
  \emph{EACL'12}.

\bibitem{kolditz19}
T.~Kolditz, C.~Lohr, J.~Hellrich, L.~Modersohn, B.~Betz, M.~Kiehntopf, and
  U.~Hahn, ``Annotating german clinical documents for de-identification,''
  \emph{Studies in health technology and informatics}, 2019.

\bibitem{hripcsak05}
G.~Hripcsak and A.~S. Rothschild, ``Agreement, the f-measure, and reliability
  in information retrieval,'' \emph{Journal of the American Medical Informatics
  Association}, 2005.

\bibitem{James13}
G.~James, D.~Witten, T.~Hastie, and R.~E. Tibshirani, \emph{An Introduction to
  Statistical Learning}, 2013.

\bibitem{sundararaman2019}
D.~Sundararaman, V.~Subramanian, G.~Wang, S.~Si, D.~Shen, D.~Wang, and
  L.~Carin, ``Syntax-infused transformer and bert models for machine
  translation and natural language understanding,'' 2019.

\bibitem{bird2009natural}
S.~Bird, E.~Klein, and E.~Loper, \emph{Natural language processing with Python:
  analyzing text with the natural language toolkit}, 2009.

\bibitem{bojanowski16}
P.~Bojanowski, E.~Grave, A.~Joulin, and T.~Mikolov, ``Enriching word vectors
  with subword information,'' \emph{CoRR}, vol. abs/1607.04606, 2016.

\end{thebibliography}

\end{document}